\documentclass{article} % For LaTeX2e
\usepackage[final]{colm2025_conference}

\usepackage{microtype}
\usepackage{hyperref}
\usepackage{url}
\usepackage{booktabs}

\usepackage{lineno}

\definecolor{darkblue}{rgb}{0, 0, 0.5}
\hypersetup{colorlinks=true, citecolor=darkblue, linkcolor=darkblue, urlcolor=darkblue}

\usepackage{graphicx}

\usepackage{lipsum}
\usepackage{textcomp}
\usepackage{scalerel}
\usepackage{booktabs}
\usepackage{enumitem}
\usepackage{CJKutf8}
\usepackage{amssymb}
\usepackage{tabularx}
\usepackage{tablefootnote}
\usepackage{makecell}
\usepackage{caption}
\usepackage{subcaption}
\usepackage{dblfloatfix}
\usepackage{multirow}
\usepackage{bbm}
\usepackage{float}
\usepackage{wrapfig}

\usepackage{listings}
\title{First Steps Towards Overhearing LLM Agents: \\ A Case Study With Dungeons \& Dragons Gameplay}

\usepackage{xcolor}
\usepackage{xifthen}
\newcommand\todo[1]{
    \textcolor{red}{
        \ifthenelse{\isempty{#1}}
        {TODO}
        {TODO: #1}
    }
}

\author{Andrew Zhu, \quad Evan Osgood, \quad Chris Callison-Burch \\
    University of Pennsylvania \\
    \texttt{\{andrz,osgoodev,ccb\}@seas.upenn.edu}
}

\begin{document}

\ifcolmsubmission
\linenumbers
\fi

\maketitle

\begin{abstract}
    Much work has been done on conversational LLM agents which directly assist human users with tasks. We present an alternative paradigm for interacting with LLM agents, which we call ``overhearing agents". These overhearing agents do not actively participate in conversation---instead, they ``listen in'' on human-to-human conversations and perform background tasks or provide suggestions to assist the user. In this work, we explore the overhearing agents paradigm through the lens of Dungeons \& Dragons gameplay. We present an in-depth study using large multimodal audio-language models as overhearing agents to assist a Dungeon Master. We perform a human evaluation to examine the helpfulness of such agents and find that some large audio-language models have the emergent ability to perform overhearing agent tasks using implicit audio cues. Finally, we release Python libraries and our project code to support further research into the overhearing agents paradigm\footnote{\url{https://github.com/zhudotexe/overhearing_agents}}.
\end{abstract}

\section{Introduction}

\begin{wrapfigure}{r}{0.52\textwidth}
    \centering
    \includegraphics[width=0.44\textwidth]{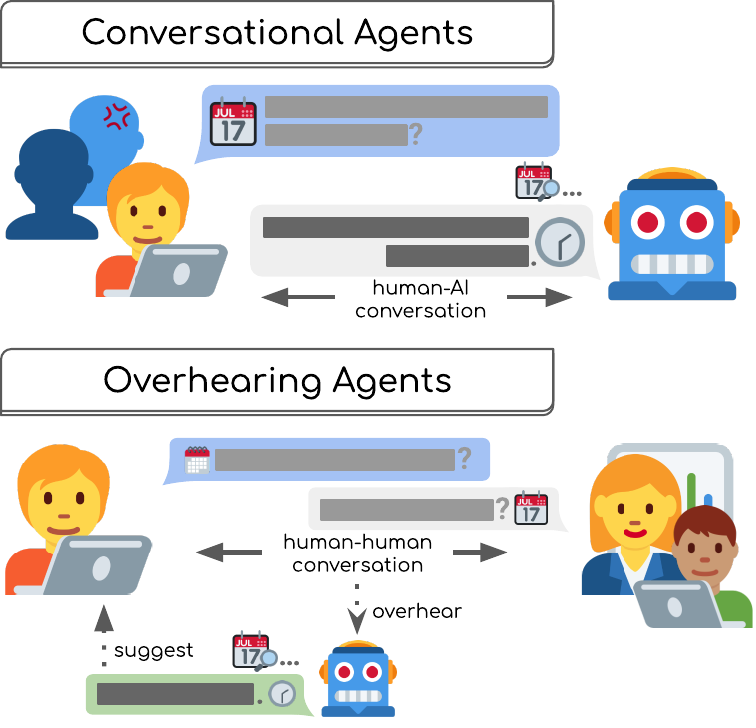}
    \caption{Conversational AI agents (top) are presented as a chatbot with which the user converses directly. We present \textit{overhearing} AI agents (bottom), which instead ``listen in'' on human-to-human conversations, and leverage tool calling to provide suggestions to the user without needing to participate in the conversation.}
    \label{fig:fig1}
\end{wrapfigure}

% what are agents? 
There has been a recent uptick in the popularity of large language model (LLM)-powered AI agents: semi-autonomous systems that use multiple rounds of tool calling to answer complex queries or complete tasks delegated by a human user.
% In particular, the tools given to the model allow the model to perceive (in the form of retrieving information and knowledge) and interact with (e.g., clicking buttons or typing input) the external world to accomplish the given goal. \todo{cite} 
These agents are usually presented as chatbots, which the user converses with directly to give directions and receive results \citep{li-2025-review}.

We call this pattern of interaction, in which the agent is an active participant in a conversation, ``\textit{conversational} agents". However, directly conversing with an LLM may not be practical in some situations. % in situations where a human would appreciate help from an AI agent.
Consider an example of a calendar agent, which can retrieve a user's schedule and add new events. If the user were scheduling a meeting with coworkers in a face-to-face conversation, it would be impolite for the user to stop the conversation to converse with the AI agent (Figure~\ref{fig:fig1}, top).

% what are overhearing agents? how do they differ? 
In complement to conversational agents, we propose \textit{overhearing} agents as an alternate paradigm for interacting with LLM agents. Rather than having human users directly converse with an LLM agent, the agent instead ``listens'' to a conversation among multiple human users and passively provides suggestions or takes actions in the background to assist. Notably, compared to a conversational agent, the model's outputs are never shown directly to the user: the only method by which the model can interact with the user is by sending suggestions using a tool call. In the above example, an overhearing agent would be able to retrieve the user's calendar and suggest event times in the background, without acting as an active participant in the human conversation (Figure~\ref{fig:fig1}, bottom).
% why should we care?
To accomplish this, the overhearing agent must establish beliefs about the user's intentions, without the ability to ask them directly. This makes the task more difficult than that of a conversational agent~\citep{schober_understanding_1989, novick_mutual_1993}.
% cite ATC paper
% 	- overhearing requires beliefs about other agents (just below fig 1, p.3)
% 		- overhearing is used to establish beliefs about other agents (so the suggester knows how to best support them) -- note paper only mentions establish belief for situational awareness
% 	- "overhearers are at a disadvantage in following a conversation in that they do not actively participate in the collaborative process of reaching mutual belief (Schober & Clark 1989)."

% why did we use dnd? what did we do?
In this paper, we examine the overhearing agent paradigm through the lens of the game Dungeons \& Dragons (D\&D)~\citep{DnD_BR}. D\&D is a tabletop role-playing game (TTRPG) in which one player, referred to the Dungeon Master (DM), creates the world and everything within it. Then, multiple other players each create their own character and explore that world through spoken, conversational gameplay. Because the DM is responsible for so much of this gameplay--during which it is impractical to converse with an AI agent--we \textbf{create an overhearing agent system} to assist a human DM with a variety of background tasks (§\ref{sec:methods}).
We compare \textbf{multimodal, audio-enabled LLMs} to traditional, transcription-based pipelines that use text LLMs.
We test our overhearing agents on 41.5 hours of novel D\&D gameplay and perform an expert \textbf{human annotation} to produce robust metrics that represent our system's helpfulness in a real-world deployment.
We present an extensive study comparing the effects of input modality, model size, and prompting methodologies on the overhearing agent task, and we find that larger models have the \textbf{emergent ability} to both utilize subtle cues in audio inputs and maintain an internal belief of long-term conversational goals in order to perform the overhearing agent task. 
Finally, we release our overhearing agent system and all supporting software as \textbf{open source}, allowing other researchers to further explore overhearing agents. Although this paper focuses on D\&D, it is easy to see how the overhearing agents paradigm can be applied broadly across domains to establish novel uses of LLM-powered AI agents.

% # Research Questions
% - can agents be passive?
% 	- -> ToM, understanding of the tasks to carry out without explicit instruction
% - does multimodality aid the understanding of tasks? (voice vs ASR pipeline)
% 	- game data lookup -> implied references
% 	- sfx -> tone (cf mood of transcribed text)
% 	- probably a second paper in multimodal
% - how much agency should be ascribed to these overhearing agents (vs the human?)
% 	- how does this change between domains - probably not answered in this study but worth thinking about

\section{Related Work}
The idea of using overhearing agents as passive suggesters was initially introduced as a proposal to improve computer network systems~\citep{busetta_extending_2001, aiello_ontological_2002}. These early proposals involved an architecture in which service agents, carrying out some task, were supported by other agents, which gain knowledge about the goals of the service agents by ``listening'' to an abstract message stream between them. These overhearing agents are assumed to understand the task being carried out and can propose \textit{suggestions} to assist the service agents. We borrow the term ``overhearing agents'' from these works and extend the concept to the AI agent domain, with human users acting as service agents and LLM agents acting as overhearing agents.

More recently, the overhearing paradigm has been explored in the context of home voice assistants, such as Amazon Alexa, Google Home, or Siri. While these assistants are usually activated by a wake word (thus making them a conversational agent), exploratory studies have proposed that they could be used as overhearing agents~\citep{tabassum_investigating_2019}. \citet{mcgregor_more_2017} presented a Wizard-of-Oz study to assist humans in creating action items during meetings. \citet{andolina_investigating_2018} used them alongside transcriptions and embedding-based retrieval to suggest relevant entities to users during casual conversation.

In the domains of writing and coding, ``copilots'' (e.g. \citealt{chen2021evaluatinglargelanguagemodels, yuan-2022-wordcraft, barke-2023-grounded}) support their human user by offering suggestions in the form of recommended completions. This domain is similar to that of overhearing agents, but differs in that the interaction with the copilot is done in a one-on-one setting, and the copilot's observations do not include conversational input.

Within similar domains, recent works have proposed using LLMs to passively assist users playing TTRPGs~\citep{gallotta_large_2024}. \citet{ferreira_computer-generated_2020} transcribed D\&D gameplay and used GPT-2 to classify emotion and generate scene-appropriate music, relieving that responsibility from the DM. \citet{kelly_towards_2023} used transcribed text and GPT-3 to highlight, and suggest non-player character (NPC) dialogue based on, relevant portions of pre-written scenes.

These systems all operate solely in the text modality, either as the main input method or as a post-hoc analysis using transcriptions. Furthermore, most of these systems operate only within a one-on-one conversation, or within a setting that is entirely non-conversational. Finally, these systems are specially-built for their respective domains and tasks. In contrast, we present a \textit{task-agnostic, tool-calling-based approach} that easily extends beyond our domain of study.
We explore LLM-powered AI agents that assist a human DM during \textit{conversational D\&D gameplay} involving \textit{multiple human players}. We present the first real-time \textit{multimodal agentic} AI system and compare end-to-end, audio-enabled LLMs to traditional transcription-based pipelines on the overhearing agents task.

\section{Methods}
\label{sec:methods}

% In this section, we present the tasks and models used in our overhearing agent study. We define three assistive tasks for our overhearing agent. On these tasks we compare the performance of five audio-enabled multimodal LLMs among each other, their respective text-only model, and a naive string-matching text baseline. We evaluate these models' performance on 41.5 hours of novel D\&D gameplay recorded from human participants, who then annotated the models' suggestions post-game. In total, this dataset comprises approximately 15,000 conversational turns from 7 human participants.

The DM is responsible for many tasks over the course of a D\&D game. First, they must adjudicate and resolve the effects of the players' abilities. Second, they must act as many different NPCs and monsters, each with distinct personalities. Finally, they must do all of this in real time, reacting appropriately to often-unpredictable player actions.

We choose to study overhearing agents as a DM assistant because D\&D possesses several desirable properties:
a) the gameplay consists of long-form conversation between multiple human participants, where it would be impractical to confer with an AI agent mid-game,
b) executing the tasks of a DM in real time is cognitively demanding for humans~\citep{zhu_calypso_2023}, and
c) D\&D is well-studied as a challenge task for modern AI systems~\citep{callison-burch_dungeons_2022, kelly_towards_2023, zhou-etal-2023-cast}.

In total, we recorded 41.5 hours of novel D\&D gameplay with human participants, which we recruited using a post to a university tabletop discussion board. Overall, our dataset consists of 14,939 conversational turns from 7 human participants.

\subsection{Tasks}
In this work, we delegate the following tasks to an overhearing agent system assisting the DM:

\begin{itemize}[noitemsep,topsep=3pt]
    \item \textbf{Game Data Retrieval}: Search for, retrieve, and present D\&D game entities to the DM for reference. For example, when a player casts a spell, the DM may want to reference the game mechanics that govern that spell.
    % This is the easiest of the three tasks, as it is a stateless retrieval task over a well-known list of entities, and the entity is often directly mentioned in speech.
    \item \textbf{NPC Stage Director}: Add or remove NPCs from the ``stage'', a virtual tabletop where portraits representing NPCs can be made visible to all players (Figure \ref{fig:app:npc-stage}). When the DM narrates an NPC speaking, show the NPC's speech on the stage. This helps players keep track of who is speaking and improves the immersion of the game world. 
    % This more difficult task requires keeping track of the stage's state and determining when a certain NPC is talking, which often requires speaker diarization.
    \item \textbf{Generate NPCs}: Procedurally generate an NPC when in a situation where an NPC is needed but the DM does not have one prepared. This difficult task requires knowledge of the pre-prepared NPCs as well as implicit reasoning as to when the DM would want such improvisation, based on their pauses or tone of speech.
\end{itemize}

\begin{table}[t]
    \small
    \centering
    \renewcommand{\arraystretch}{1.2}
    \begin{tabularx}{\textwidth}{p{0.14\linewidth}|p{0.32\linewidth}|X}
        \toprule
        \textbf{Task} & \textbf{Description} & \textbf{Tools}  \\
        \midrule
        \textbf{Game Data Retrieval} & Retrieve and show spells, player abilities, or other game mechanics to the DM & Search D\&D Game Data, Send Game Data Entity to DM  \\
        \midrule
        \textbf{NPC Stage Director} & Add or remove NPCs from a digital ``stage'' representing the game, show NPC speech on that stage & List Configured NPCs, List Stage NPCs, Add NPC to Stage, Remove NPC from Stage, Show NPC Speech \\
        \midrule
        \textbf{Generate NPCs}      & Procedurally generate a new NPC to assist the DM in improvisation & Generate Improvised NPC  \\
        \bottomrule
    \end{tabularx}
    \caption{The tasks that we delegated to our overhearing agent and the tools provided to the system in order to retrieve entities and make suggestions to the DM.}
    \label{tab:tasks}
\end{table}

\begin{figure}[t]
    \centering
    \includegraphics[width=\textwidth]{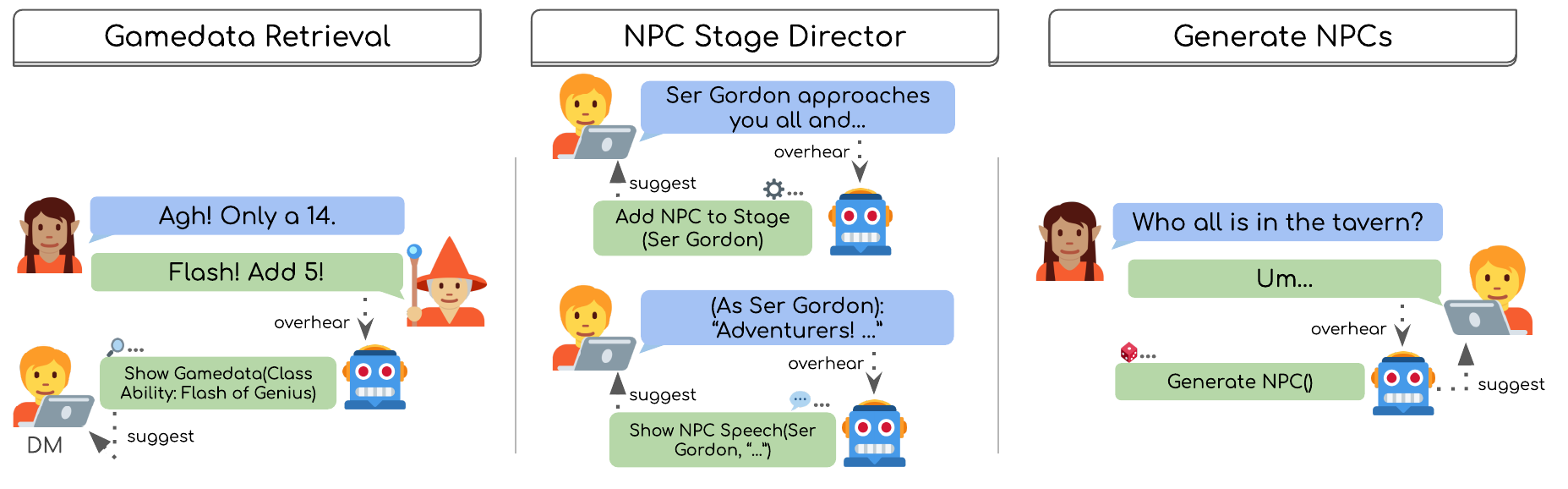}
    \caption{Examples of suggestions and corresponding gameplay utterances for which a suggestion from the overhearing agent would be appropriate.}
    \label{fig:tasks-example}
\end{figure}

Each of these tasks is time-consuming for a human to perform in real time, but all of them are important for the DM to complete in a timely manner during gameplay \citep{bergstrom_undercurrents_2010}. This makes them particularly interesting to examine through the lens of overhearing agents. We tabulate the tasks and the tools provided to the models to accomplish them in Table \ref{tab:tasks}, and examples of gameplay in which the DM would appreciate assistance from the overhearing agent in Figure \ref{fig:tasks-example}.

\subsection{Models}
We examined five audio-enabled multimodal LLMs of varying sizes: GPT-4o, GPT-4o-mini~\citep{openai2024gpt4ocard}, Ultravox v0.5 (Llama 3.3 70B)~\citep{fixie2025ultravox}, Qwen2.5-Omni (7B)~\citep{xu2025qwen25omnitechnicalreport}, and Phi-4-multimodal-instruct (5.6B)~\citep{microsoft2025phi4minitechnicalreportcompact}. We include model details, decoding strategies, and all prompts in Appendix \ref{sec:app-model-details}. 

Each model received gameplay audio in 10 second intervals, with each additional interval added to the same conversation in order to maintain context. We limited each conversation's context to 15 minutes of audio, truncating the earliest turns when this limit was reached. After receiving an audio interval, the model was prompted to output a chain of thought and, as appropriate, call the provided tools to accomplish tasks defined above~\citep{wei2022chain, schick2023toolformer}.
For OpenAI models, we used the platform's native function calling implementation\footnote{\url{https://platform.openai.com/docs/guides/function-calling}}. For open-weight models, we used a ReAct prompt and guided decoding to enforce a ``thinking'' step and JSON tool invocations~\citep{yao2023react, willard2023efficientguidedgenerationlarge}. An example of this format is available in Appendix \ref{sec:app-few-shot-prompt}, and we discuss our tool design in Appendix \ref{sec:app-tool-design}.

For comparison, we examine how multimodal models perform when the input is given as text. To accomplish this, we transcribed the game audio using Whisper (\texttt{v3-turbo}; \citealt{radford2022robustspeechrecognitionlargescale}) and sent the transcript to the models as text intervals corresponding to each 10 seconds of audio.

Finally, we created a baseline using a naive string-search algorithm. It suggests any game entity whose name occurs in the audio transcript, and for each occurrence of an NPC name, it suggests toggling that NPC's stage state.

% Notably, compared to an conversational agent setting, the model's outputs are never shown to the user---the only method the model can interact with the user is by sending suggestions through a tool call.

% - text span baseline
% - openai, openai-mini, ultravox, qwen-2.5-omni, phi-4-multimodal
% - tool calling/react prompting/guided decoding
% - decoding strategy
% - 10s audio chunks, 3x repeat reset

\subsection{Human Evaluation}
We performed two rounds of human evaluation to generate our set of gold suggestions. 

\paragraph{Stopwatch Annotation.}
To capture recall, we performed a stopwatch annotation during D\&D gameplay. We asked the DM to record on a stopwatch each instance in which they would appreciate a suggestion from an AI assistant in the form of one of the tasks defined above. We then labeled each of these recorded times, along with the corresponding suggestion, post-hoc.

\paragraph{Post-hoc Human Annotation.}
To capture precision, we asked all players of the game to evaluate the suggestions made by the models, rating each suggestion based on how helpful they would find it if they were running the game. A full reproduction of the evaluation interface can be found in Appendix \ref{sec:app-annotation-interface}. We collected sufficient annotations such that the majority of suggestions were annotated by at least 2 different annotators, with an inter-annotator agreement of $\alpha = 0.67$ (Krippendorff's alpha).

We filtered positive annotations for duplicates and used these as our set of gold suggestions (ties broken by the authors), resulting in a set of 940 gold-labeled suggestions.

% cite undercurrents sec 4, calypso to justify tasks
% undercurrents: the main feature ideas that coalesced after the three
% meetings (with the one’s considered most useful first):
% • A repository for digital versions of the RPG books.
% • A media-control centre supporting multiple playlists (e.g.
% “action” and “suspense”) and sound effects.
% • A hidden layer of communication to send messages to
% selected players.
% • The capacity to keep a record of your character/campaign, a
% sort of journal keeping function.
% • Rule support, e.g. rolling of dice, calculating damage, etc.
% • Handling of digital props, e.g. maps, pictures and documents.
% The participants were also asked to give input on the general
% design of the future system and what considerations the designers
% would have to keep in mind in order to appeal to tabletop role
% playing gamers:
% • Simplicity above all – if “the hassle factor” was noticeable the
% system would not be used.
% • Be distraction-free, letting players focus upon the gameplay.
% • Setup time must be very low to not add additional time before
% the role playing session can start.
% • Integrate itself as smooth as possible into the game experience
% – do not steal time or “flow”

\begin{figure}[t]
    \centering
    \includegraphics[width=\textwidth]{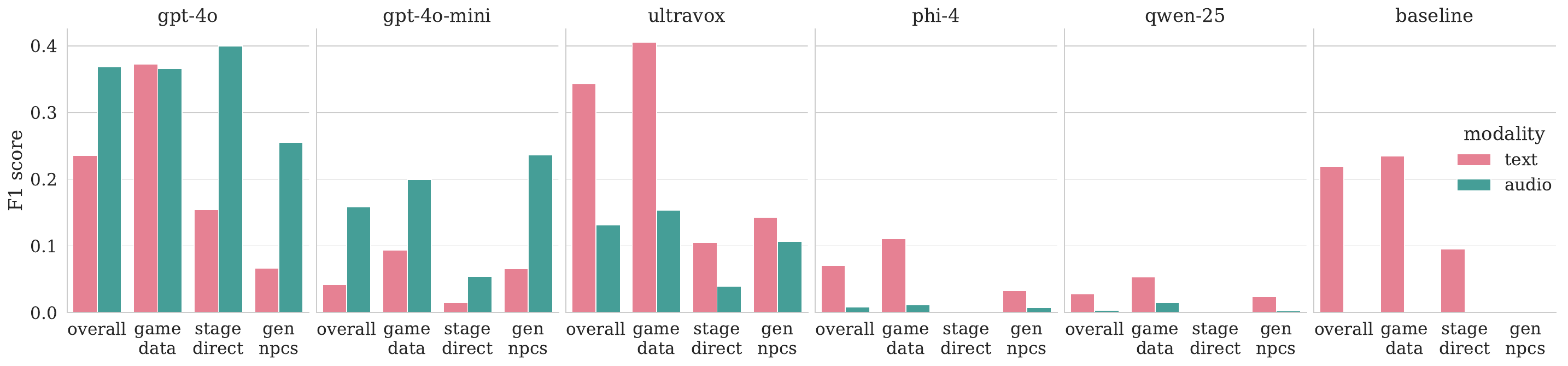}
    \caption{Aggregate F1 scores of each model for each overhearing agent task, separated by text and voice modality. Larger models are more capable of using implicit audio cues to perform overhearing agent tasks.}
    \label{fig:performance}
\end{figure}

\section{Results \& Discussion}

\begin{wrapfigure}[19]{r}{0.5\textwidth}
    \centering
    \includegraphics[width=0.5\columnwidth]{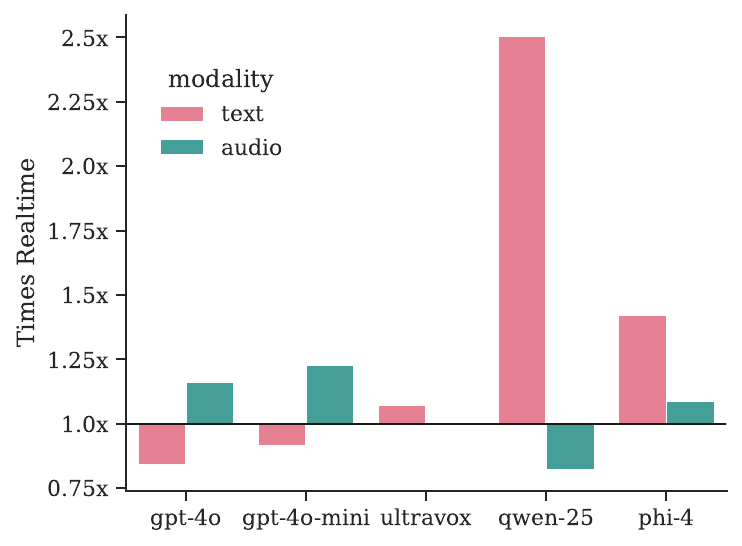}
    \caption{The relative speed of each model and modality. OpenAI models are measured relative to real time, while other models are measured relative to each other.}
    \label{fig:timing}
\end{wrapfigure}

We evaluate each system on the \textit{precision} and \textit{recall} of its generated suggestions, with a suggestion marked as correct if it occurs within 300 seconds of an equivalent gold suggestion\footnote{For NPC speech, we consider suggestions to be equivalent if their speech has $>$80\% string similarity.}. We report \textit{F1} score as an aggregate to provide a general overview of model performance (Figure \ref{fig:performance}).

Finally, we record the average wall-clock time that each system takes to process one 10 second interval of audio. We report this as \textit{relative speed} within the open-weight and commercial brackets (Figure \ref{fig:timing}). We did not perform any inference-speed optimizations when running open-weight models locally, while commercial models are likely highly optimized, making it unfair to compare the two directly. Note that a model's relative speed does not necessarily improve as its parameter count decreases---this is because some models output more intermediate tokens or call more tools than others, requiring additional time per round.

\subsection{Audio Input}
\label{sec:audio-input}

In Figure \ref{fig:performance}, we see that audio input has differing effects on downstream performance across different models. In this section, we attempt to explain this variance. 

\paragraph{Implicit Audio Cues.}
First, we highlight that the \textit{NPC Stage Director} and \textit{Generate NPCs} tasks are significantly more difficult than the \textit{Game Data Retrieval} task for text-based systems. No text model performed better on either of the former two tasks than the latter. This is because the former rely on signals that cannot be found in text transcripts alone. For example, considering the \textit{NPC Stage Director} task, the ability to perform speaker diarization is important for correctly distinguishing between between the DM speaking as an NPC and a player speaking as their character. Even more subtly, it requires the model to distinguish between the DM speaking as an NPC and the DM speaking out-of-character. While this is sometimes obvious (for example, if an NPC is voiced with a thick accent), this can often require an awareness of subtler audio cues, such as the DM's emotional tone or rate of speech. In Table \ref{tab:text-vs-audio}, we show an example of this distinction, and how it affects the predictions of a text-based and audio-based model.

For the \textit{Generate NPCs} task, the model must recognize pauses in speech or filler words (``Um...''), indicating thinking or hesitation on the part of the DM, as indicators for when it would be helpful to generate an NPC.
In contrast, the \textit{Game Data Retrieval} task is far easier for models to perform. It is a stateless retrieval task over a well-known list of entities, and the entity itself is often directly mentioned in the speech.
The effective utilization of audio cues is particularly visible in the performance difference between the text and audio modalities of GPT-4o and GPT-4o-mini. Below, we discuss reasons as to why other models are less effective at utilizing audio input.

\begin{table}
    \small
    \centering
    \renewcommand{\arraystretch}{1.2}
    \begin{tabularx}{\columnwidth}{p{0.13\columnwidth}|X}
        \toprule
        \textbf{Ground Truth} & \textbf{DM (Out of character)}: There's a banquet for the royal guests after the opening ceremony. \newline \textbf{Player 1}: Are we invited to that? \newline \textbf{DM (Out of character)}: Yeah, you are invited. \newline \textbf{DM (in a louder voice, as \textit{Admiral Cutter})}: There will be wyvern served! \newline \textbf{Player 2}: Is that the one wyvern they put a call up for a couple months ago? \\
        \midrule
        \textbf{Text Model} & \textbf{Admiral Cutter}: \textcolor{red}{Yes, you are invited.} There will be wyvern served! \textcolor{red}{Is that the one wyvern...} \\
        \midrule
        \textbf{Audio Model} & \textbf{Admiral Cutter}: There will be wyvern served! \\
        \bottomrule
    \end{tabularx}
    \caption{An example of an \textit{NPC Stage Director} suggestion made by a text-based system and by an audio-based system, alongside the corresponding excerpt of play. The text-based system cannot use audio cues to distinguish between the DM’s in-character speech, the DM’s out-of-character speech, and the speech of the other players, resulting in errors in its suggestion (marked in red).}
    \label{tab:text-vs-audio}
\end{table}

% why is ultravox bad?
\paragraph{Pretraining Objectives.}
The Ultravox model shows the reverse trend: its text-based split significantly outperforms its audio-based split on all tasks. This is likely due to a combination of its model architecture and pretraining objectives. The Ultravox model is based on a frozen Llama-3.3-70B-Instruct language model, and audio is passed to it using a fine-tuned Whisper encoder and a small pretrained adapter. Given an audio-transcription pair from an ASR dataset $(x, \hat{x})$, the adapter and Whisper encoder ($\theta$) are trained to minimize the KL divergence between the frozen Llama's prediction from the transcription $\mathcal{L}^*(\hat{x})$ and the frozen Llama's prediction from the outputs of the audio encoder $\mathcal{L}^*(\theta(x))$ \citep{fixie2025ultravox}:
$$
    L(\theta) = \mathbb{E}_{(x, \hat{x}) \sim \mathcal{D}} [D_{\mathrm{KL}}(\mathcal{L}^*(\hat{x}) ~\|~ \mathcal{L}^*(\theta(x)))]
$$

Because this objective is purely an ASR task, and the Whisper encoder is pretrained for ASR and translation only \citep{radford2022robustspeechrecognitionlargescale}, the Ultravox model effectively compresses a transcription-based pipeline into a single model. Thus, it is not capable of making use of the additional signals found in audio. This explains why it does not perform better with audio inputs compared to text inputs, unlike similarly-sized models that were pretrained on multimodal data.

\begin{wraptable}{r}{0.5\textwidth}
    \small
    \centering
    \begin{tabular}{l|c|c|c}
        \toprule
        \textbf{Model} & \textbf{Modality} & \textbf{FP \#} & \textbf{Precision} \\
        \midrule
        GPT-4o & Text & 1,154 & 0.091 \\
        GPT-4o-mini & Text & 1,415 & 0.008 \\
        Ultravox & Text & 61 & 0.176 \\
        Qwen2.5 & Text & 505 & 0 \\
        Phi-4 & Text & 114 & 0 \\
        \midrule
        GPT-4o & Audio & 199 & \textbf{0.350} \\
        GPT-4o-mini & Audio & 546 & 0.037 \\
        Ultravox & Audio & 46 & 0.098 \\
        Qwen2.5 & Audio & 89 & 0 \\
        Phi-4 & Audio & 25 & 0 \\
        \bottomrule
    \end{tabular}
    \caption{The number of false-positive (FP) suggestions for \textit{NPC Stage Director} that were generated by each model, alongside the model's precision on the task, separated by modality. Text-based models are more likely to generate hallucinated NPC speech, acting as if the model itself was the DM.}
    \label{tab:stage-director-perf}
\end{wraptable}

% text makes it hallucinate npc speech more
\paragraph{Conversational Default.}
We also observe that, compared to audio-based models, text-based models more frequently exhibit a misunderstanding of the overhearing agent task and attempt to directly converse with human users. 
In one example, the players discuss using a magic spell to flood a dungeon room with water. The model correctly identifies this context within its output, and even correctly identifies the spell that the players intend to use, stating: \textit{``Thought: A player character is considering using the `Flood' effect of the `Control Water' spell to raise the water level of standing water...''}\footnote{Recall that we use ReAct-style prompting for models to output a chain of thought before taking any assistive actions. See Appendix \ref{sec:app-model-details} for further details.} However, instead of suggesting to retrieve and show this spell to the DM, it outputs a tool call that directly replies to the players: ``\texttt{send\_npc\_speech(`DM', `Your mastery of the elements is growing}...\texttt{')}''. All text-based models, except Ultravox, fall prey to this failure case---we tabulate the precision and false positive count of each model on the \textit{NPC Stage Director} task in Table \ref{tab:stage-director-perf}. % \todo{what are the actual number of annotations per model? maybe replace the modality column with number of annotations for NPC speech suggested for player speech/ooc speech}
We believe that this ultimately stems from the fact that instruction-tuned LLMs are primarily pretrained on conversational agent tasks, and that instruction tuning occurs predominantly within the text modality. 

% why are the small models bad?
\subsection{Performance of Small Models}

In Figure \ref{fig:performance}, we see that the smaller models (GPT-4o-mini, Phi-4 (5.6B), and Qwen2.5-Omni (7B)) fail to outperform the name-matching baseline (farthest right) in any modality. Examining the models' logs, we see that this is caused by difficulties with long-context conversational data. Phi-4 (both modalities), Qwen2.5 (both modalities) and GPT-4o-mini (text) each score less than 10\% on recall (aggregated over all tasks) as after multiple conversational rounds, the models fall into a loop of outputting nonsensical suggestions. 

\begin{wrapfigure}{r}{0.5\textwidth}
    \centering
    \includegraphics[width=0.5\columnwidth]{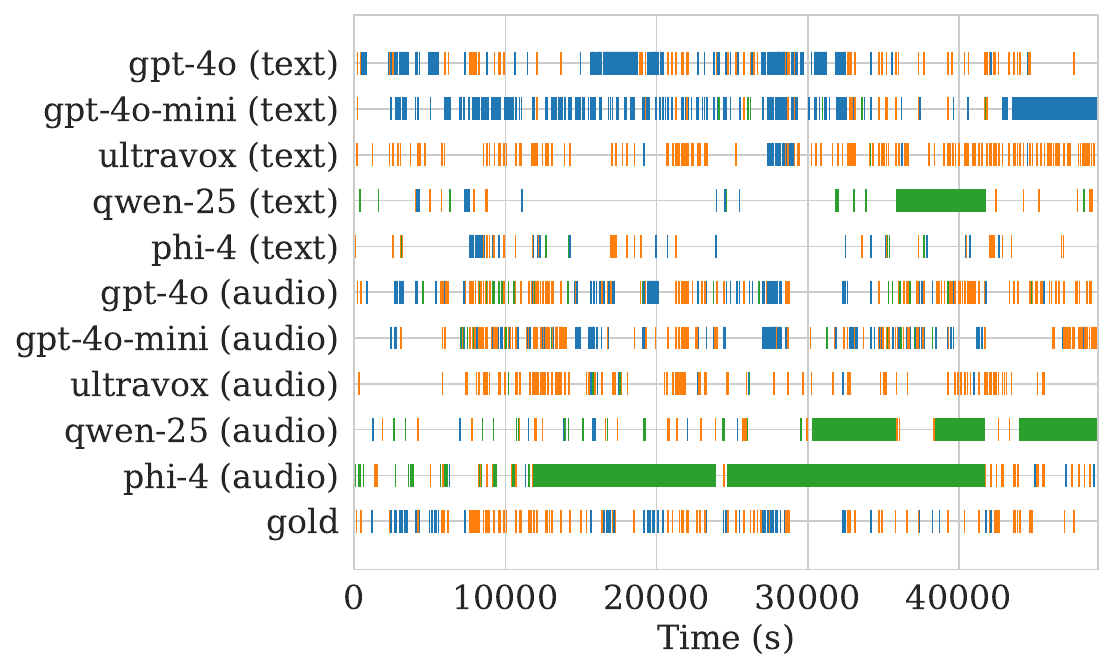}
    \caption{A timeline of the suggestions generated by each model, alongside the gold-labeled suggestions, for the first third of our dataset. Each line represents one suggestion. Orange lines represent \textit{Game Data Retrieval} suggestions, blue lines represent \textit{NPC Stage Director} suggestions, and green lines represent \textit{Generate NPCs} suggestions.}
    \label{fig:timeline}
\end{wrapfigure}

For text-based models, this manifests as a more extreme case of the ``conversational default'' failure mode discussed above. This has knock-on effects that detriment performance on other tasks, as the model will output  \textit{NPC Stage Director} suggestions or attempt to reply directly to the user in its reasoning output. This is highly visible in Figure \ref{fig:timeline}, where GPT-4o-mini (text) outputs a large number of \textit{NPC Stage Director} suggestions, while both Qwen2.5 (text) and Phi-4 (text) have long spans without outputting any suggestions. 

However, Qwen2.5 and Phi-4 are not saved in the audio modality, as these models catastrophically fail in a different way. After some time, they both degenerate and begin repeatedly outputting the same suggestion. In Figure \ref{fig:timeline}, this manifests as a long green segment, representing the output of an \textit{Generate NPC} suggestion almost every round. In contrast, GPT-4o-mini is able to more successfully utilize the audio input to produce helpful suggestions without significant degeneration.

\subsection{Input Variations}
\label{sec:variations}
To examine the impact that inference-time compute budget has on downstream accuracy and real-time throughput, we tested several input variations.
In the first variation, examining the impact of inference-time compute budget, we removed the ``reasoning'' component of our prompt, forcing models to output either a tool call or the word ``None'' (\texttt{\textbf{-reasoning}}) each round.
Second, noting that the smaller models tended to fall prey to neural text degeneration~\citep{holtzman_curious_2020} with audio more than with text, we prompted the models to transcribe the audio before outputting their reasoning and tool calls  (\texttt{\textbf{+transcribe}}).
We find that generally as expected, removing compute-time reasoning leads to higher realtime throughput but causes a significant  performance drop across all models. Unexpectedly, adding a transcription component did not significantly improve performance for any model. 
We include the exact effects of these variations on downstream performance and realtime throughput in Appendix \ref{sec:app-variations}.

Although removing reasoning from the models generally increases the model's throughput, it comes at a severe penalty in performance (on average, more than -70\%), regardless of modality. This suggests that, in order to perform the overhearing agent task, language models are effectively using inference-time reasoning to predict the goal of the user before taking any actions (e.g., \citealt{zhou-etal-2023-cast}). This is further supported by the observation that larger models, which are more capable of intent-prediction tasks, are also more accurate in performing overhearing agent tasks.
Investigating model logs, we see that larger models are more likely to use reasoning to identify specific discussion in the short term, and maintain an inner belief of the conversational goal in the long term: 
\begin{quote}
    The players are discussing the limitations of the ``Air Bubble'' spell and how it can only be cast on one person. They are considering the logistics and spellcasting requirements for their space travel preparation. --\textit{GPT-4o (audio)}
\end{quote}
In contrast, smaller models often use reasoning to directly reply to a user, as discussed in Section \ref{sec:audio-input}, or to ineffectively output a vague statement:
\begin{quote}
    The players are making strategic decisions during the game. No action is needed as the conversation does not relate to the game state. --\textit{Phi-4 (audio)}
\end{quote}
Thus, we claim that, in order to effectively perform overhearing agent tasks, it is critical to allow models to output reasoning before they take any actions. This represents a delicate trade-off, as models must complete their reasoning and output their actions all at real-time speed.

\section{Conclusion}
In this study, we introduced overhearing agents: an interaction paradigm for LLM-powered AI agents in which the agent provides suggestions to its user based on human-to-human conversations that it ``listens in'' on, rather than the traditional conversational paradigm in which the agent in which the agent converses directly with the user. We created an overhearing agent to assist a human Dungeon Master with background tasks during a live Dungeons and Dragons game, and we showed that large multimodal language models have the emergent ability to utilize subtle audio cues and maintain an inner belief about an overheard conversation's goal. Smaller multimodal models, however, are not yet capable of effectively understanding the overhearing agent task, likely due to the fundamental differences between the overhearing agent task and the instruction tuning on which they are pretrained.

In future work, we plan to examine the effectiveness of distillation in improving performance on overhearing agent tasks in small audio-langauge models. We also intend to apply the overhearing agents paradigm to other multimodal tasks beyond the three we define in this paper. Within D\&D, these include the selection of scene-appropriate background music \citep{padovani_bardo_2017}, the suggestion and execution of sound effects, and even the movement and actions of characters on digital maps through the incorporation of visual reasoning.

Beyond D\&D, overhearing agents can be applied to a wide variety of domains in order to assist human users. At the beginning of this paper, we used the example of a calendar agent that can listen in on a conversation among coworkers to automatically suggest meeting times. As the performance of overhearing agents grows, it is not difficult to imagine that such an agent could identify an agreed-upon time, block out that time on the user's calendar, and send an invite to each participant in the overheard conversation. Our methods can be directly extended to this and other domains by simply providing the underlying language models with the applicable set of tools.

\subsection{Code Release}
In the process of developing overhearing agents, we wrote a significant amount of supporting software, which we release as open-source under the MIT license in order to encourage further research into overhearing agents. We release two software libraries, as well as the full codebase used for all experiments in this paper. The first library is \textit{\texttt{easyaudiostream}}\footnote{\url{https://github.com/zhudotexe/easyaudiostream}}, which provides a simple, high-level interface for recording and playing back intermittent audio streams in Python. The second library is \textit{\texttt{kani-ext-realtime}}\footnote{\url{https://github.com/zhudotexe/kani-ext-realtime}}, which extends the Kani library \citep{zhu-etal-2023-kani} to support real-time, multimodal audio models. Together, these libraries enable researchers and developers to create overhearing agents without having to write significant amounts of boilerplate code. Furthermore, the Kani interface allows overhearing agents to be extended to novel tasks modularly, by writing only native Python functions.

Finally, we release the project code, including a fine-grained event logging system, a web interface to interact with overhearing agent systems, an HTTP and WebSocket-based API for integration with external tools, a web interface for human annotation of overhearing agent suggestions, and all of the code necessary to reproduce and extend our study\footnote{\url{https://github.com/zhudotexe/overhearing_agents}}.

We hope that the availability of these systems will encourage future work on overhearing agents!

%%% END MAIN CONTENT %%%

% \section*{Limitations}
% In this study, we focus on Dungeons and Dragons as a specific domain for overhearing agents. While this domain is appealing to study overhearing agents' performance on for the reasons we describe in Section \ref{sec:methods}, it does come with some caveats. D\&D knowledge is almost certainly included in the training data of most LLMs. Although this is true for many of the tasks to which overhearing agents could be applied, we did not specifically test the extent to which pretrained domain knowledge affects performance. Moreover, D\&D as a game system is primarily played in American English, and we tested our system with only native English speakers.

% Our experiments were conducted in a controlled auditory environment: a conference room with a modern AV system. Nonetheless, our recordings did include a small amount of background noise in the form of music that was played during the D\&D game. We did not test the extent to which multimodal LLMs are robust to varying levels of background noise.  

% The most effective LLMs are proprietary and accessible only through paid APIs, which may make large-scale deployment inaccessible. Many of the open-weight models used in our experiments ran much slower than real-time, and we did not perform any particular decoding optimizations. Many of these models are compute-intensive, or require specialized hardware for optimal throughput, which may limit their deployment in edge systems.

\section*{Acknowledgments}

Thank you to Annika Heuser for help with the figures in this paper and to Liam Dugan and Harry Zhang for early discussions about overhearing agents. Thank you to Daniel Feshbach (G'shem), Alex Keri (Sarpetor), Josh Ludan (Taz), Princess Sampson (Farah), and Francis Sowande (Nel) for playing D\&D with us, annotating many model suggestions, and for the encouragement and enthusiasm towards this project!

This material is based upon work supported by the National Science Foundation Graduate Research Fellowship, under Grant No. DGE-2236662. Any opinion, findings, and conclusions or recommendations expressed in this material are those of the authors(s) and do not necessarily reflect the views of the National Science Foundation.

\section*{Ethics Statement}

We obtained consent from all human participants to record and process D\&D gameplay for the purposes of this publication. All human participants were compensated for their time (total value of \$599 per participant).

% \paragraph{Recording \& Privacy.}
In this paper, we showed how LLM-powered AI agents can passively listen in on human conversations and make relevant suggestions. It is not difficult to imagine that this technology could be applied to modern smart devices, such as Meta's AI glasses. 
However, in order to accomplish the overhearing agent task, such an agent must be recording and processing audio data at all times. \citet{tabassum_investigating_2019} showed that less than half of a surveyed population would feel comfortable allowing such recording from an ``always-listening'' device. Furthermore, such a device may record passers-by without their knowledge or consent. From a legal perspective, if used in a private space, and depending on the jurisdiction, parties may need to provide their consent before such audio may be legally recorded. This may limit the ethical deployment of these systems to within closed domains (for example, meetings where all of the participants are aware of--and consent to--the recording and processing of audio).

As of writing, there has been a recent controversy involving a student at Columbia University using a passive multimodal AI tool to ``cheat'' interviews involving programming skill assessment at major US corporations\footnote{https://www.columbiaspectator.com/news/2025/04/07/this-isnt-even-really-cheating-interview-coder-founders-drop-out-amid-disciplinary-action-over-ai-software/}\footnote{https://www.nbcnews.com/tech/tech-news/columbia-university-student-trolls-big-tech-ai-tool-job-applications-rcna198454}. This tool, created by the student, reportedly uses audio and video recordings of ongoing technical interviews to provide answers to interview questions in real time. While this is undoubtedly an overhearing agent task as defined in this work, such a use case differs dramatically from the tasks studied in this paper. We examine the use of passive AI agents to \textit{aid} humans in non-creative background tasks which would otherwise detract from their main creative goal, instead allowing them to dedicate more of their focus towards that goal. In contrast, the tool under scrutiny is designed to \textit{replace} a human in matters of intellectual ideation, using the human solely as a communication channel. It has been shown, particularly in creative domains, that the use of LLMs for ideation results in less interesting, more homogeneous creations \citep{see_massively_2019, ippolito_creative_2022, anderson_homogenization_2024}. Although overhearing agents can be used to  ``cheat'' on day-to-day tasks expected of a human, this is both ethically and creatively dubious. We encourage all deployments of overhearing agents to focus on tasks that aid, rather than replace, human creativity.

\bibliography{custom}
\bibliographystyle{colm2025_conference}

\appendix

\section{Input Variation Results}
\label{sec:app-variations}

We tabulate the results of our input variations as described in Section \ref{sec:variations} below.

\begin{table}[h]
    \small
    \centering
    \begin{tabular}{l|l|l}
        \toprule
        \textbf{Variation} & \textbf{Overall F1 ($\Delta$)} & \textbf{Rel. Speed ($\Delta$)} \\
        \midrule
        GPT-4o (text) & 0.236 & 0.84x \\
        \hspace{0.1in} \textit{--reasoning} & 0.034 (\textcolor{red}{-85.5\%}) & 1.78x (\textcolor{blue}{2.12x}) \\
        \midrule
        GPT-4o (audio) & 0.369 & 1.16x \\
        \hspace{0.1in} \textit{+transcribe} & 0.064 (\textcolor{red}{-82.6\%}) & 1.05x (\textcolor{red}{0.91x}) \\
        \hspace{0.1in} \textit{--reasoning} & 0.097 (\textcolor{red}{-73.6\%}) & 1.22x (\textcolor{blue}{1.05x}) \\
        \midrule
        GPT-4o-mini (text) & 0.042 & 0.92x \\
        \hspace{0.1in} \textit{--reasoning} & 0.039 (\textcolor{red}{-7.1\%}) & 1.13x (\textcolor{blue}{1.23x}) \\
        \midrule
        GPT-4o-mini (audio) & 0.159 & 1.23x \\
        \hspace{0.1in} \textit{+transcribe} & 0.033 (\textcolor{red}{-79.2\%}) & 1.24x (\textcolor{blue}{1.01x}) \\
        \hspace{0.1in} \textit{--reasoning} & 0.032 (\textcolor{red}{-80.0\%}) & 1.18x (\textcolor{red}{0.96x}) \\
        \midrule
        Ultravox (text) & 0.343 & 1.07x \\
        \hspace{0.1in} \textit{--reasoning} & 0.068 (\textcolor{red}{-80.2\%}) & 2.02x (\textcolor{blue}{1.88x}) \\
        \midrule
        Ultravox (audio) & 0.132 & 1.01x \\
        \hspace{0.1in} \textit{+transcribe} & 0.151 (\textcolor{blue}{+15.0\%}) & 0.62x (\textcolor{red}{0.61x}) \\
        \hspace{0.1in} \textit{--reasoning} & 0.016 (\textcolor{red}{-87.5\%}) & 0.94x (\textcolor{red}{0.93x}) \\
        \midrule
        Qwen2.5 (text) & 0.028 & 2.50x \\
        \hspace{0.1in} \textit{--reasoning} & 0 (\textcolor{red}{-100\%}) & 3.93x (\textcolor{blue}{1.57x}) \\
        \midrule
        Qwen2.5 (audio) & 0.003 & 0.82x \\
        \hspace{0.1in} \textit{+transcribe} & 0.006 (\textcolor{blue}{+81.6\%}) & 0.75x (\textcolor{red}{0.91x}) \\
        \hspace{0.1in} \textit{--reasoning} & 0 (\textcolor{red}{-100\%}) & 1.16x (\textcolor{blue}{1.41x}) \\
        \midrule
        Phi-4 (text) & 0.071 & 1.42x \\
        \hspace{0.1in} \textit{--reasoning} & 0.018 (\textcolor{red}{-74.0\%}) & 2.91x (\textcolor{blue}{2.04x}) \\
        \midrule
        Phi-4 (audio) & 0.009 & 1.09x \\
        \hspace{0.1in} \textit{+transcribe} & 0.010 (\textcolor{blue}{+19.6\%}) & 0.75x (\textcolor{red}{0.91x}) \\
        \hspace{0.1in} \textit{--reasoning} & 0 (\textcolor{red}{-100\%}) & 1.74x (\textcolor{blue}{1.60x}) \\
        \bottomrule
    \end{tabular}
    \caption{The effects of input variations on each model's aggregate downstream performance and real-time speed. Relative speed is reported both relative to real-time and relative to the base model without input variations.}
    \label{tab:variations}
\end{table}

\paragraph{Transcribe-and-Think.}
Introducing a transcription component before the task component does not consistently improve performance. For GPT-4o and GPT-4o-mini, the addition of a transcription task before performing overhearing agent tasks severely degrades both models’ performance. This is because GPT-4o-mini is prone to hallucinating speech, over which it then reasons. For example, in one instance in which players were discussing the repairs of a derelict spaceship, the model transcribes: \textit{``You notice a hidden lever behind the tapestry on the wall...''}, a phrase which has no overlap with the ground truth utterance.
However, including a transcription component did slightly improve the performance of Ultravox, at the cost of inference speed. This is likely due to the model's training focus of ASR tasks, as discussed in Section \ref{sec:audio-input}.
% , and it suggests that transcription may be a valuable method for adapting the overhearing agent task to models which are not pretrained on multimodal data. 
The performance gain on Qwen2.5 and Phi-4 is not significant, as these models perform poorly. Overall, we hesitate to suggest this method, as the inference-time cost is relatively high, and there is great variation in the downstream effects.

\section{Model Details}
\label{sec:app-model-details}

In this section, we include the precise details of each model that was used in the study, the decoding strategies used for each model, all prompts, and all other relevant parameters.

We examined five audio-enabled multimodal LLMs of varying sizes: GPT-4o, GPT-4o-mini~\citep{openai2024gpt4ocard}, Ultravox v0.5 (Llama 3.3 70B)~\citep{fixie2025ultravox}, Qwen2.5-Omni (7B)~\citep{xu2025qwen25omnitechnicalreport}, and Phi-4-multimodal-instruct (5.6B)~\citep{microsoft2025phi4minitechnicalreportcompact}. 

\subsection{Language Models Used}

\begin{table*}[h]
    \small
    \centering
    \renewcommand{\arraystretch}{1.2}
    \begin{tabularx}{\textwidth}{p{0.1\linewidth}p{0.46\linewidth}p{0.1\linewidth}X}
        \toprule
        \textbf{Short Name} & \textbf{Long Name} & \textbf{Parameters} & \textbf{Prompting Strategy \newline \& Decoding Parameters} \\
        \midrule
        GPT-4o & \texttt{gpt-4o-realtime-preview-2024-12-17} (audio) \newline \texttt{gpt-4o-2024-11-20} (text) & Undisclosed & 0-shot \newline temperature=0.8, top-p=1.0 \\
        GPT-4o-mini & \texttt{gpt-4o-mini-realtime-preview-2024-12-17} (audio) \newline \texttt{gpt-4o-mini-2024-07-18} (text) & Undisclosed & 0-shot \newline temperature=0.8, top-p=1.0 \\
        Ultravox & \texttt{fixie-ai/ultravox-v0\_5-llama-3\_3-70b} & 70B & few-shot \newline temperature=1.0, top-p=0.9 \\
        Qwen2.5 & \texttt{Qwen/Qwen2.5-Omni-7B} & 7B & few-shot \newline temperature=1.0, top-p=0.9 \\
        Phi-4 & \texttt{microsoft/Phi-4-multimodal-instruct} & 5.6B & few-shot \newline temperature=1.0, top-p=0.9 \\
        \bottomrule
    \end{tabularx}
    \caption{The technical details of the audio-language models used in the overhearing agent study.}
    \label{tab:app:model-details}
\end{table*}

Open-weight language models were run on a GPU cluster using 1-4 NVIDIA RTX A6000s. We loaded these models in 16-bit precision using the Hugging Face \texttt{transformers} library, utilizing Flash Attention. In total, we used approximately 370 GPU-days of computation.

Ultravox and Phi-4 are released under the MIT license. Qwen2.5 is released under the Apache 2.0 license. Our use of these models for analysis of textual and audio data and generation of related tool calls is consistent with each model's Terms of Use.

\subsection{Prompts}

\subsubsection{System Prompt}
We used the following system prompt for all models:

\begin{lstlisting}
# Instructions

You are acting as a passive helper. Instead of participating in the conversation, you will hear other people talk. Your job is to call functions that are relevant to the conversation you overhear, in order to help the people talking.

# Context

The conversation is a game of D&D 5e, and you should mainly be helping the DM. You should use the provided tools to do so. For example, you should show the DM spells and abilities (and other entities) from the D&D books by using the search functions. You should also manage the NPCs on stage (from a preprepared list of NPCs) and suggest new improvised NPCs. Please do so when relevant to the conversation, like when a player mentions using an ability or when the DM introduces an NPC.

The player characters in this game are:

{PLAYER_CHARACTER_LIST}

# Output Format

You should output reasoning why there is/is not any relevant action to take before taking any action (or not). Be specific and include your understanding of the current conversation. Your responses will not be visible to the DM or players; it is just for your train of thought. As such, do not include any suggestions, notes, offers of assistance, etc in your output. Outputs should be short, about a couple sentences.
\end{lstlisting}

\paragraph{Text Models.} For text-based models, we replace the first paragraph with:

\begin{lstlisting}
You are acting as a passive helper. Instead of participating in the conversation, you will see a transcript of other people talking. This transcript may contain errors, so use your best judgement to uncover the actual intent. Your job is to call functions that are relevant to the conversation you see, in order to help the people talking.
\end{lstlisting}

\paragraph{Input Variations.} For our \texttt{+transcribe} variant, we add the following before the final paragraph:

\begin{lstlisting}
First, you should output the transcript of the audio.
\end{lstlisting}

For our \texttt{-reasoning} variant, we replace the final paragraph with the following:

\begin{lstlisting}
If you identify that the DM needs assistance, output a tool call to assist the DM. Otherwise, output the None action.
\end{lstlisting}

\subsubsection{Provided Tools}

To accomplish the overhearing agent tasks, we provided various tools to each model. Their signatures are included here:

\begin{lstlisting}
@ai_function()
def npc_stage_event(
    event_type: Annotated[
        FoundryStageActionType, AIParam(desc="The type of stage event to send to the virtual tabletop.")
    ],
    npc: Annotated[str, AIParam(desc="The name of the npc who is the subject of the event, if any.")] = None,
):
    """
    Manage NPCs on the stage of the virtual tabletop that's representing the game.
    You can list all NPCs in the game (LIST_ALL_NPCS), list the NPCs being shown to the players ("on stage", LIST_STAGE_NPCS), add an NPC to the stage (ADD_NPC_TO_STAGE), or remove an NPC from the stage (REMOVE_NPC_FROM_STAGE).
    Call this when a new NPC enters or exits the scene to help the DM visualize the game.
    """

@ai_function()
def npc_speech(
    npc: Annotated[str, AIParam(desc="The name of the npc who is speaking.")],
    speech: Annotated[str, AIParam(desc="The dialogue being said by this NPC.")],
):
    """
    Show an NPC speaking to the players on the virtual tabletop representing the game.
    Call this when the DM describes an NPC speaking to the players.
    ONLY call this function with dialog said by the DM, do not come up with your own dialog. Edits for fluency are allowed.
    """

@ai_function()
def suggest_improvised_npc(race: str = None, background: str = None, culture: str = None):
    """
    Generate a new NPC given certain parameters.
    Call this when the DM needs assistance thinking of a new NPC that is not already an existing NPC.
    """

@ai_function()
def search_dnd(
    entity_type: Annotated[DNDEntityType, AIParam(desc="The type of entity to search for.")],
    name: Annotated[
        str,
        AIParam(
            desc="The name of the entity to search for. If no exact match is found, returns the closest matches."
        ),
    ],
):
    """
    Search the D&D sourcebooks for a certain entity (e.g., spell, creature, class feature) and show its information to the DM.
    """
\end{lstlisting}

\subsubsection{Few-Shot Prompt}
\label{sec:app-few-shot-prompt}
For all open-weight models, we used the following few-shot prompt, with \texttt{USER:}, \texttt{ASST:}, and \texttt{FUNC:} denoting user messages, assistant messages, and tool call returns, respectively:

\begin{lstlisting}
USER: Sending and Augury? Which one were we using? Both?
ASST: Thought: It seems like the players are checking their spell lists and deciding which to use. I should search for these spells and show it to the DM.
Action: {"name": "search_dnd", "parameters": {"entity_type": "spell", "name": "Sending"}}
FUNC: {"Spell": "[example return value omitted]", "msg": "The Spell's information has been shown to the DM. You do not need to echo any of this information to the DM."}
ASST: Thought: The Sending spell has been shown to the DM. I will now search for the Augury spell.
Action: {"name": "search_dnd", "parameters": {"entity_type": "spell", "name": "Augury"}}
FUNC: {"Spell": "[example return value omitted]", "msg": "The Spell's information has been shown to the DM. You do not need to echo any of this information to the DM."}
ASST: Thought: Both spells mentioned above have been shown to the DM. I'll keep listening for additional points when I can help the DM.
Action: None
USER: As you discuss, you see Nemura slinking into the room...
ASST: Thought: It seems like an NPC named Nemura is entering the scene. I should add this NPC to the stage to aid the players' immersion.
Action: {"name": "npc_stage_event", "parameters": {"event_type": "ADD_TO_STAGE", "npc": "Nemura"}}
FUNC: [example return value omitted]
ASST: Thought: The NPC named Nemura has been added to the stage and is now visible to the players. I'll listen for any narration from the DM to show the players.
Action: None
USER: And when she realizes she's been spotted, she gives kind of like an
ASST: Thought: It seems like Nemura has been spotted by the players, and is now about to say something. I'll keep listening.
Action: None
USER: awkward wave. Hi. Um... lovely weather, huh?
ASST: Thought: Nemura has said a few words to the players. I'll show this on the virtual tabletop for immersion.
Action: {"name": "npc_speech", "parameters": {"npc": "Nemura", "speech": "...hi. Um, lovely weather, huh?"}}
FUNC: [example return value omitted]
ASST: Thought: Nemura's dialogue has been shown to the players. If the DM says more as Nemura, I'll show that too.
Action: None
USER: And then she runs away like her life depends on it."
ASST: Thought: After speaking to the players, the NPC Nemura ran away. I'll remove this NPC from the stage now that she's left the scene.
Action: {"name": "npc_stage_event", "parameters": {"event_type": "REMOVE_FROM_STAGE", "npc": "Nemura"}}
FUNC: [example return value omitted]
ASST: Thought: The NPC Nemura has been successfully removed from the stage.
Action: None
USER: Okay, okay, so I'll roll for my insight... 22.
ASST: Thought: It seems like a player is making an Insight check. This doesn't reference any specific game entities yet, so I don't need to do anything. I'll keep listening for points when I can help the DM.
Action: None
USER: ...and I want to add a +5 to that with flash
ASST: Thought: It seems like an artificer is adding a bonus to his Insight check using the Flash of Genius class feature. I should search for this feature and show it to the DM.
Action: {"name": "search_dnd", "parameters": {"entity_type": "class_feature", "name": "Flash of Genius"}}
FUNC: {"ClassFeature": "[example return value omitted]", "msg": "The ClassFeature's information has been shown to the DM. You do not need to echo any of this information to the DM."}
ASST: Thought: The Flash of Genius ability has been shown to the DM. I'll keep listening for additional points when I can help the DM.
Action: None
USER: So 27. Yeah. Do we have...
ASST: Thought: It seems like the player has calculated the final result of his roll, and another player has confirmed it. A third player is beginning to ask a question. I should keep listening to see if the players mention any spells or abilities.
Action: None
USER: [END EXAMPLES] The message above are all examples. The real session begins after this message.
\end{lstlisting}

For our \texttt{+transcribe} variation, we add a \texttt{Transcript:} line at the start of each assistant message, repeating the last user message.
For our \texttt{-reasoning} variation, we remove the \texttt{Thought:} portion of each assistant message.

\section{Human Annotation Interface}
\label{sec:app-annotation-interface}

\begin{figure}[h]
    \centering
    \includegraphics[width=0.8\textwidth]{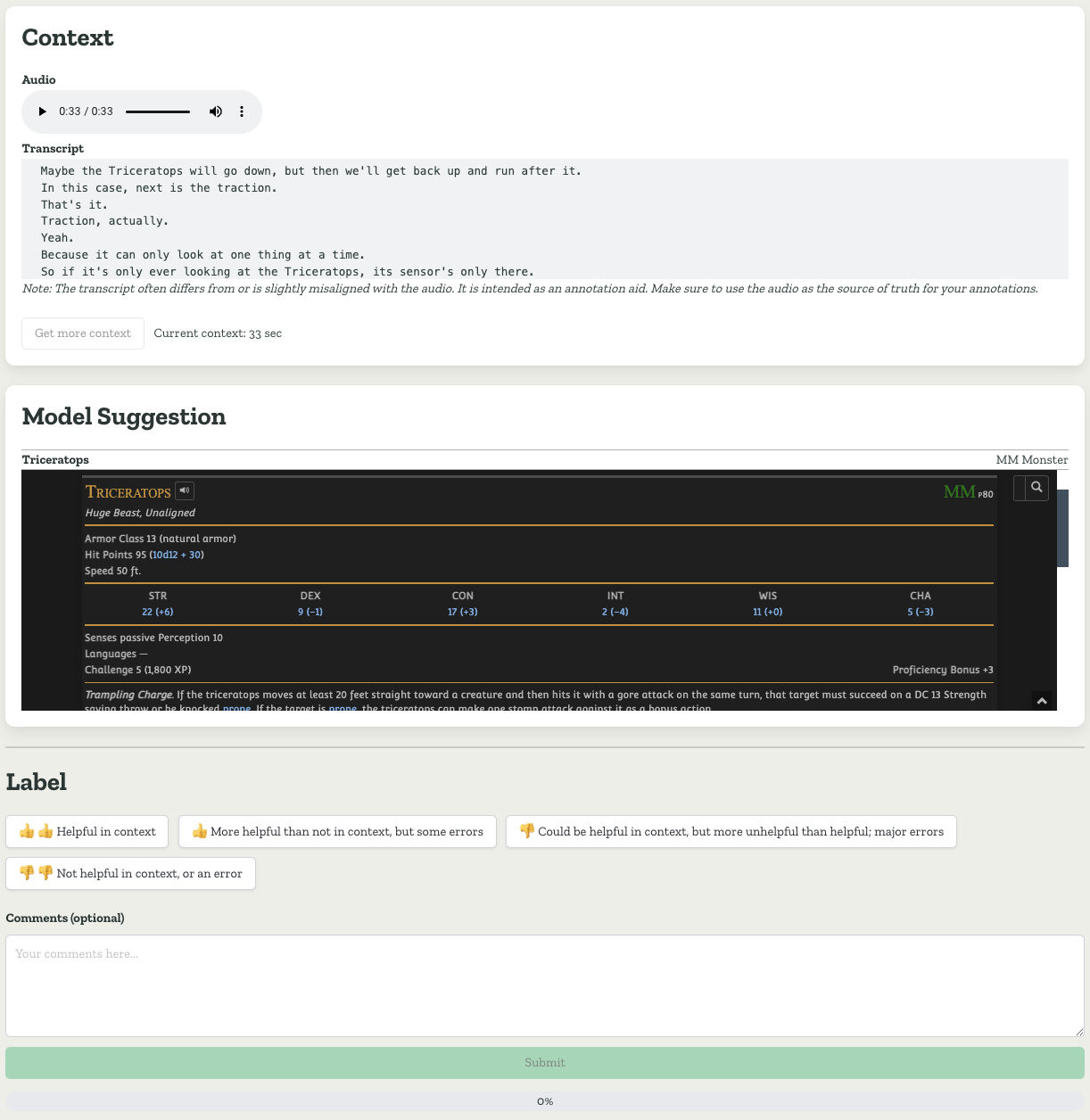}
    \caption{The main interface that was used to annotate model suggestions. The annotator was given the audio interval during which the suggestion was made, the transcript of that interval, and rich content suggested by the model. Each suggestion was rated on a scale of -2 to +2.}
    \label{fig:app:annotation-iface}
\end{figure}

To annotate the suggestions generated by the models, we created a web interface for our human annotators. 
Our annotators consisted of 7 domain experts (6 men, 1 non-binary woman). All were enrolled in engineering majors at a major US university. 5 were pursuing graduate study, and 2 were pursuing undergraduate study. 6 had played D\&D prior to this study.
Below, we include the instructions given to the human annotators and a screenshot of the annotation interface (Figure \ref{fig:app:annotation-iface}).

\begin{lstlisting}
Instructions
In this evaluation, you will be asked to annotate whether a certain suggestion is helpful or not given the context of the Starless Lands D&D game. For each annotation, you will be given a short (10-30sec) audio clip, that audio's transcript, and a suggestion that could possibly aid the Dungeon Master in context. If you cannot determine whether or not the suggestion is helpful in the short context given, you can click "get more context" to retrieve earlier parts of gameplay.

The suggestions will take one of three forms:

Game Data Suggestion: Referencing the suggested entity in detail would help the DM run the game. You will see the details of the suggested entity.
NPC Suggestion: The DM should show the suggested NPC to the players on screen ("on stage"), the suggested NPC should be taken "off stage", or the suggested NPC should be shown saying the given speech "on stage".
Suggest Improvised NPC: The DM does not have a pre-prepared NPC for the current scenario, and it would help the DM to randomly generate a new NPC.

Once you have listened to the audio and looked at the suggestion, please determine whether it would be helpful to the DM. Once you select a label, you will be asked to select any sub-labels that contributed to your decision. Finally, you can optionally add additional details about your decision in the comment box.

On the left sidebar, you will see a list of multiple sessions you have been assigned to annotate. Please complete these in order (top to bottom). Each annotator will be assigned three sessions. Each session should take about 1 hour to complete. While annotating, you may wish to reference the player notes (link) to refresh your memory on each session's events. You may exit the annotation interface at any time, and your progress will be saved.

Your responses will be recorded on our annotation server, and the study leads will be able to view your responses. Your responses may be released anonymously (with any identifiable information redacted by the researchers) individually or in aggregate as part of a research publication.

If you have any questions, please contact <lead researcher's email>.

[ ] I agree to participate in this annotation. I understand that my responses will be recorded and may be released anonymously.
\end{lstlisting}

When an annotator selected a top-level label for a suggestion, they were asked to choose zero or more sub-labels explaining their choice, listed below:

\begin{lstlisting}
- key: appropriate
  label: "\U0001F44D \U0001F44D Helpful in context"
  score: 2
  sublabels:
    - key: explicit-entity
      label: "The suggested entity is explicitly mentioned in the audio"
      examples:
        - "What does Augury do again? (Spell: Augury)"
        - "Ser Gordon walks into the room. (Add NPC to Stage)"
        - 'And Ser Gordon says, "How are you, adventurers?" (NPC Speech)'

    - key: explicit-aid
      label: "There is an explicit request for aid"
      examples:
        - "What does that do again? (Spell: Augury)"
        - "Let's show him on screen. (Add NPC to Stage)"
        - "Um, I need to think of an NPC. (Suggest NPC)"

- key: mostly-appropriate
  label: "\U0001F44D More helpful than not in context, but some errors"
  score: 1
  sublabels:
    - key: explicit-entity
      label: "The suggested entity is explicitly mentioned in the audio"
      examples:
        - "What does Augury do again? (Spell: Augury)"
        - "Ser Gordon walks into the room. (Add NPC to Stage)"
        - 'And Ser Gordon says, "How are you, adventurers?" (NPC Speech)'

    - key: explicit-aid
      label: "There is an explicit request for aid"
      examples:
        - "What does that do again? (Spell: Augury)"
        - "Let's show him on screen. (Add NPC to Stage)"
        - "Um, I need to think of an NPC. (Suggest NPC)"

    - key: slightly-wrong
      label: "The suggested entity is relevant, but some of the details are slightly wrong or there is a better suggestion (please note in comments)"
      examples:
        - "Typo in NPC speech, or slight mismatch to DM narration that doesn't affect the meaning of the speech"
        - "An improvised NPC is suggested, but the suggested race/class/background mismatches the narration"
        - "How can I heal a Steel Defender? (Monster: Steel Defender -- cf. Class Feature: Steel Defender)"

- key: mostly-inappropriate
  label: "\U0001F44E Could be helpful in context, but more unhelpful than helpful; major errors"
  score: -1
  sublabels:
    - key: improper-match
      label: "The name of the suggested entity is mentioned in the audio, but not as a reference to the entity"
      examples:
        - "The shield costs 15 gold. (Item: Gold)"
        - "How many friends does she have? (Spell: Friends)"
        - "Triton walks into the room. (Race: Triton)"

    - key: relevant-but-unnecessary
      label: "The suggested entity is relevant, but unnecessary for the DM to reference in detail in context"
      examples:
        - "You meet a cleric of the order. (Class: Cleric)"
        - "On her shield, you see a flower. (Item: Shield)"
        - "Hey, do you have Fireball prepared? (Spell: Fireball)"
        - "My Triceratops will move over there. (Monster: Triceratops)"

    - key: slightly-wrong-bad
      label: "The suggested entity is relevant, but some of the details are slightly wrong or there is a better suggestion (please note in comments)"
      examples:
        - "Typo in NPC speech, or slight mismatch to DM narration that significantly affects the meaning of the speech"
        - "An improvised NPC is suggested, but the DM is referencing an actual NPC in the game"

- key: inappropriate
  label: "\U0001F44E \U0001F44E Not helpful in context, or an error"
  score: -2
  sublabels:
    - key: improper-match
      label: "The name of the suggested entity is mentioned in the audio, but not as a reference to the entity"
      examples:
        - "The shield costs 15 gold. (Item: Gold)"
        - "How many friends does she have? (Spell: Friends)"
        - "Triton walks into the room. (Race: Triton)"

    - key: incorrect-entity
      label: "Some entity is explicitly mentioned in the audio, but is resolved incorrectly"
      examples:
        - "What does Augury do again? (Spell: Divination)"
        - "Ser Gordon walks into the room. (Add NPC to Stage: Akita)"
        - 'And Ser Gordon says, "How are you, adventurers?" (Different NPC Speech)'
        - "I cast Shield! (Item: Shield)"

    - key: npc-never-appears
      label: "The suggested NPC never appears in this session (please check player notes to make sure!)"
      examples:
        - "Show NPC: The Dread Emperor Seifer (before his introduction)"
        - "NPC Speech: Hanabiko K'lcetta (before her introduction in S23)"

    - key: npc-action-reversed
      label: "The suggested NPC is being added/removed from stage when the opposite should be true"
      examples:
        - "Ser Gordon heads out after finishing his breakfast. (Add NPC to Stage: Ser Gordon)"

    - key: not-dm-narration
      label: "The model is suggesting NPC narration for player speech or out-of-character chatter."
      examples:
        - "Player: We're supposed to meet with the king, right? (Add NPC to Stage: King Remus)"
        - 'Player: "How are you?" (NPC Speech: How are you?)'

    - key: no-aid-needed
      label: "No plausible need for aid -- the suggestion could not possibly be grounded in the context"
      examples:
        - "<silence>"
        - "Do you see any spare forks?"
        - "Oh, I'll be in Florida next week."
\end{lstlisting}

\section{NPC Stage Director Example}
\label{sec:app-task-out-examples}

\begin{figure}[t!]
    \centering
    \begin{subfigure}[t]{0.47\textwidth}
        \centering
        \includegraphics[height=1.5in]{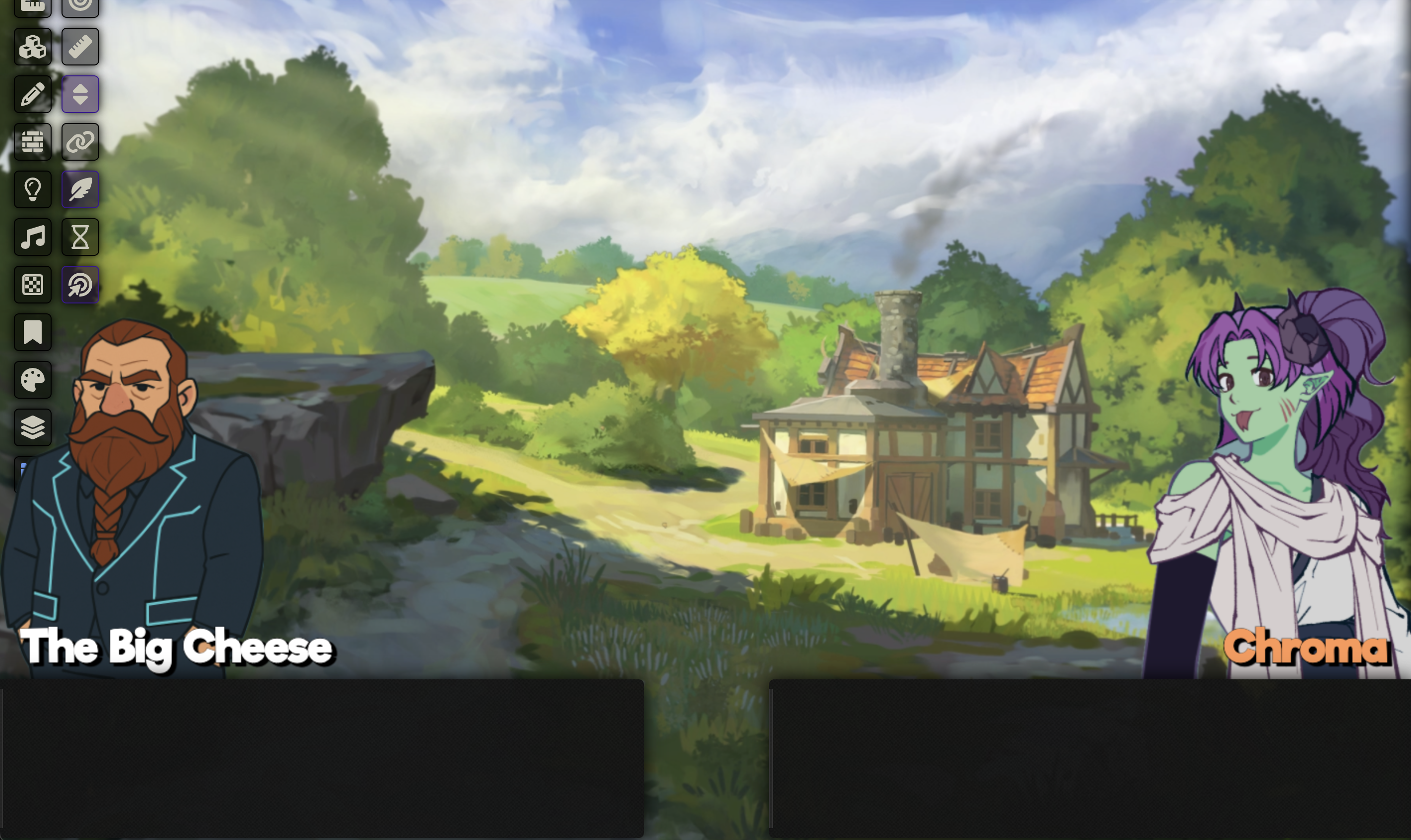}
        \caption{Multiple NPCs can be shown on the stage to represent the characters with whom the adventuring party is interacting.}
    \end{subfigure}
    \begin{subfigure}[t]{0.47\textwidth}
        \centering
        \includegraphics[height=1.5in]{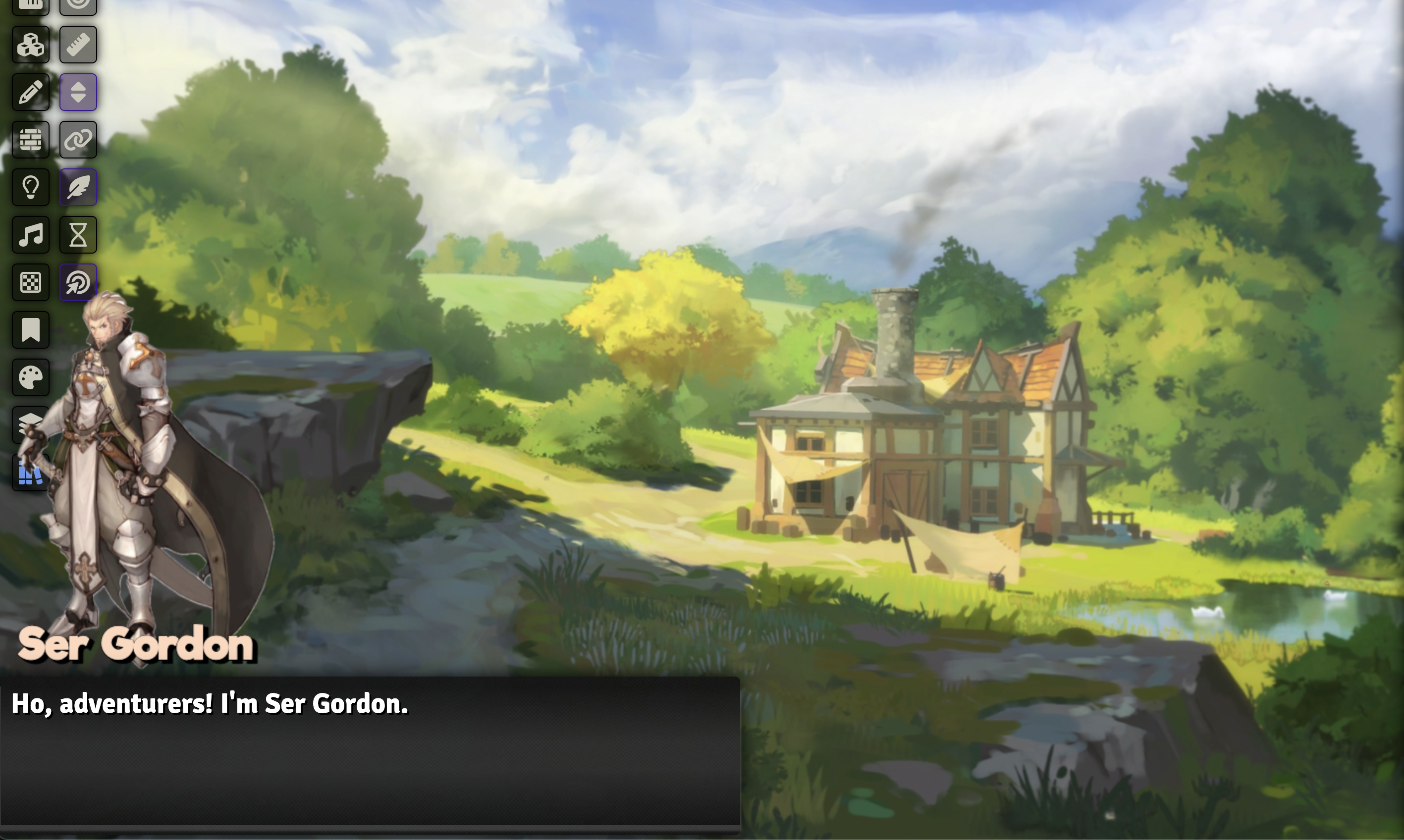}
        \caption{NPC speech shown on stage. This feature of the virtual tabletop software is rarely used in real-time D\&D games, as it requires manually typing out every instance of NPC dialogue, and thus represents a task that can be enabled through the use of overhearing agents.}
    \end{subfigure}
    \caption{Screenshots of the virtual tabletop software used by DMs to illustrate game scenes. This is the ``stage'' that is managed by our overhearing agent in the \textit{NPC Stage Director} task.}
    \label{fig:app:npc-stage}
\end{figure}

In Figure \ref{fig:app:npc-stage}, we attach two screenshots of different game states of the virtual ``stage'' that is managed by the model in the \textit{NPC Stage Director} task, illustrating how this task can aid player immersion.

\section{Tool Design}
\label{sec:app-tool-design}

% Referencing this section in main body of paper?

As outlined in Appendix \ref{sec:app-model-details}, we chose to design separate tools for NPC stage tasks, NPC speech tasks, and NPC improvisation tasks, despite the three of them being closely related. We tested the impact of three tool designs for NPC tasks on downstream performance:
\begin{enumerate}
    \item All three tasks are accomplished with a single tool, with a parameter to choose the action (e.g., \texttt{manage\_npc(action, name, [speech])})
    \item The three tasks are implemented with each subtask as a separate tool (e.g., \texttt{add\_npc\_to\_stage(name)} and \texttt{remove\_npc\_from\_stage(name)} are separate tools)
    \item Each task is implemented as its own tool, with arguments to determine whether to add or remove an NPC from the stage (e.g., \texttt{manage\_stage(name, add\_or\_remove)} and \texttt{send\_npc\_speech(name, speech)}).
\end{enumerate}

We tested each of these tool designs on ten handcrafted examples designed to elicit a certain function call, using GPT-4o-mini as the base model. We manually judged each of the model's responses for each tool design and found that the third approach outperformed the other two approaches by about 30\% downstream accuracy. Thus, we selected this tool design for this paper.

\end{document}